\documentclass[10pt,twocolumn,letterpaper]{article}

\usepackage[pagenumbers]{cvpr}
\usepackage{adjustbox}
\usepackage{multirow}
\usepackage{cuted}

\definecolor{tabfirst}{RGB}{178,34,34}
\definecolor{tabsecond}{RGB}{210,125,20}
\newcommand{\fst}[1]{\textcolor{tabfirst}{#1}}
\newcommand{\snd}[1]{\textcolor{tabsecond}{#1}}
\newcommand{\method}{CoTL}
\newcommand{\vggto}{VGGT-$\Omega$}

\newcommand{\gs}{3D Gaussian}

\newcommand{\retenk}{RealEstate10K}

\newcommand{\psnr}{PSNR}
\newcommand{\ssim}{SSIM}
\newcommand{\lpips}{LPIPS}

\newlength{\qualimgw}
\newlength{\qualraise}
\newlength{\genimgw}
\newcommand{\genimage}[1]{%
  \includegraphics[width=\genimgw]{#1}%
}
\newcommand{\genhead}[1]{%
  \scriptsize\strut #1%
}

\newlength{\qualhimgw}
\newcommand{\qualhimage}[1]{%
  \includegraphics[width=\qualhimgw]{#1}%
}
\newcommand{\qualhlabel}[1]{%
  \small\strut #1%
}
\newcommand{\qualhgroup}[1]{%
  \small #1%
}
 
\definecolor{cvprblue}{rgb}{0.21,0.49,0.74}
\usepackage[pagebackref,breaklinks,colorlinks,allcolors=cvprblue]{hyperref}

\def\paperID{} %
\def\confName{\xspace}
\def\confYear{\xspace}

\title{CoTL: Compact Token Learning for 3D Scene Representation}

\author{
Han Liang\,\textsuperscript{1,2*\dag} \quad Sibo Wu\,\textsuperscript{3*} \quad
Anpei Chen\,\textsuperscript{3\dag} \quad R\"udiger Westermann\,\textsuperscript{1,2}\\[0.35em]
\textsuperscript{1}\,TU Munich \qquad
\textsuperscript{2}\,MCML \qquad
\textsuperscript{3}\,Westlake University
}

\begin{document}
\maketitle
\begin{strip}
  \centering
  \includegraphics[width=0.97\textwidth]{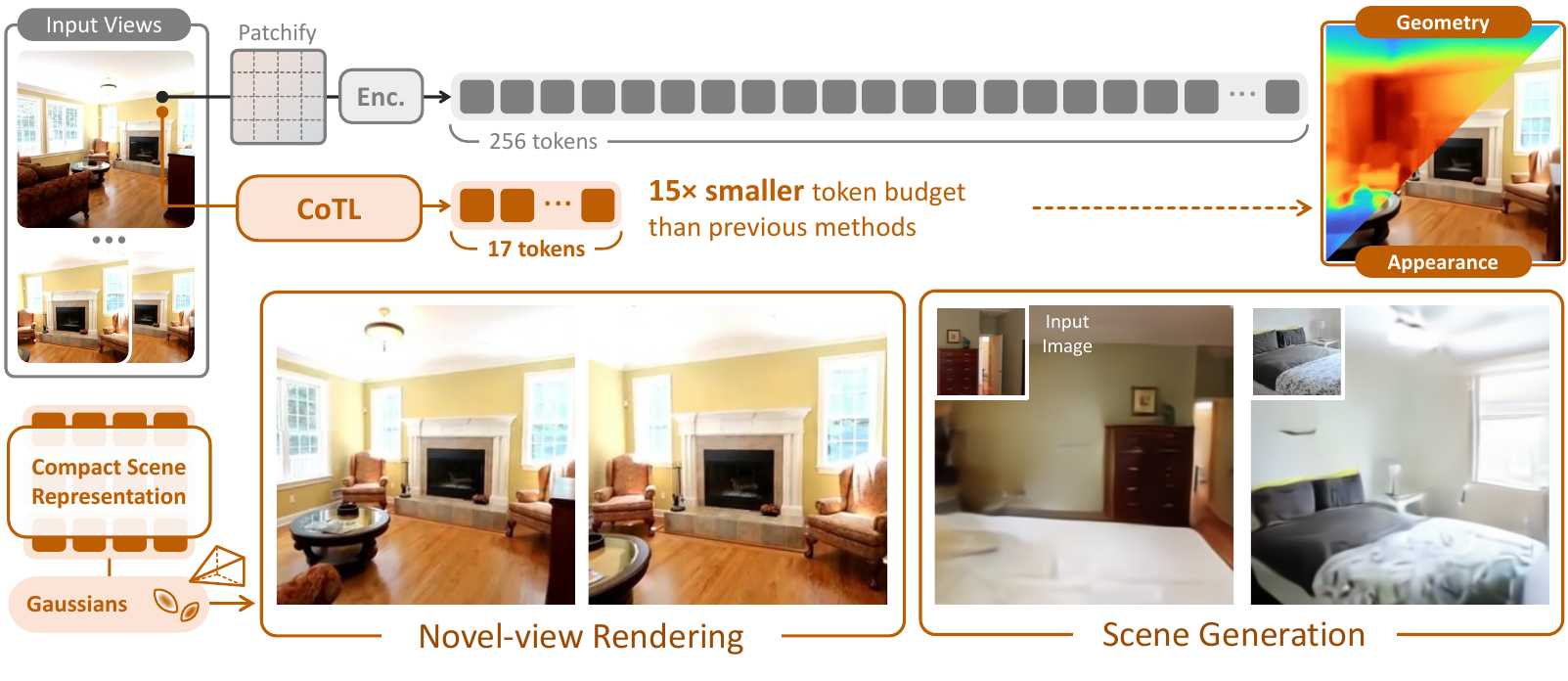}

  \captionsetup{hypcap=false}
  \captionof{figure}{\textbf{Compact token learning for 3D scene
  representation.}
  \method{} compresses dense per-view image features into a compact scene representation with 17 tokens per view, reducing the token budget by roughly $15\times$ relative to a dense patch-token grid. A query-based decoder with scene-independent Gaussian queries accesses only these tokens to reconstruct an explicit, renderable 3D Gaussian scene. The resulting representation supports novel-view rendering and can further serve as the latent space of a generative model for 3D scene generation.}
  \captionsetup{hypcap=true}

  \label{fig:regime_example}
\end{strip}
\renewcommand{\thefootnote}{\fnsymbol{footnote}}
\footnotetext[1]{Co-first author.}
\footnotetext[2]{Co-corresponding author.}
\renewcommand{\thefootnote}{\arabic{footnote}}
\begin{abstract}
We present \method{}, a framework that learns a compact set of tokens
that preserves scene geometry and appearance from unposed multi-view
images. 
Our encoder compresses the input images into small per-view
sets of latent tokens that aggregate information across views to form a view-consistent scene
representation. 
Each additional view expands the representation with
only a small, resolution-independent set of tokens, allowing its
capacity to grow with the available observations.
To force this compact bottleneck to retain sufficient scene information, a query-based
Gaussian decoder uses scene-independent queries to attend to the learned
tokens and recover an explicit, renderable set of 3D Gaussian primitives.
Novel-view rendering supervision trains the compact representation
end to end through the reconstructed Gaussian scene.
Experiments on \retenk{} and ACID show that the resulting compact, unstructured
token set supports competitive novel-view synthesis and can serve as
the latent space of a generative model for 3D Gaussian scene generation.
Together, these results show that 3D Gaussian reconstruction is an
effective objective for compact scene representation learning and
the learned token space provides a basis for broader downstream tasks
in 3D vision.
\end{abstract}
\section{Introduction}
\label{sec:intro}
Learning a scene representation from multi-view images requires compressing view-dependent observations into a compact encoding of the underlying 3D world. 
This representation must preserve shared geometry and appearance across views while discarding redundant view-specific information. Yet many recent feed-forward reconstruction models retain dense feature
representations produced by pretrained multi-view geometry backbones,
comprising hundreds of patch tokens per image.
A dominant class of feed-forward reconstruction methods uses dense pixel-aligned features as the visual memory, 
predicting pixel-aligned Gaussian primitives from dense per-view feature grids~\cite{ye2024noposplat,ye2025yonosplat,jiang2025anysplat}.
Other approaches decouple the Gaussian output budget from the input grid using clustered visual tokens, 
global latent tokens, adaptive 3D anchors, or learned queries~\cite{veicht2026zipsplat,itkin2026globalsplat,in2026atsplat,ren2026tokengs,an2025c3g,li2026querysplat}. This indicates that the explicit scene output need not inherit the size and layout of the image grid, but the decoder still retrieves scene information from dense per-view patch features. A compact Gaussian output therefore does not imply a compact latent scene representation.

Compact scene representations provide a practical interface for downstream models. 
In 3D scene understanding, vision--language models often aggregate dense scene features into a compact set of tokens, enabling scene-level representations to be efficiently processed within the context budget of a language model~\cite{hong20233dllm,chen2024ll3da,lin2026qwen3d}.
Compact representations are particularly useful for generation because they allow scene distributions to be modeled in a lower-dimensional latent space rather than directly over high-dimensional signals~\cite{rombach2022high}.
Some existing 3D object-level generation models operate on 1D shape
tokens~\cite{zhang2023vecset,zhao2025hunyuan}, and recent work compresses
entire scenes into 1D tokens for generation or
    reconstruction~\cite{wei20263drae,asim2026scenetok,chen2026nova3r}.
In these methods, however, the tokens either encode geometry without
appearance~\cite{zhang2023vecset,zhao2025hunyuan,chen2026nova3r} or are
decoded directly into target views without yielding an explicit, renderable
3D scene~\cite{wei20263drae,asim2026scenetok}. These designs leave open how
to learn a compact scene-level representation that jointly carries geometry
and appearance and can be decoded into an explicit, renderable 3D scene.

This raises a fundamental question: can a small set of learned tokens replace the dense feature grid at the encoder--decoder interface, while preserving the geometry and appearance needed to reconstruct an explicit, renderable 3D scene?

In this work, we explicitly study compact scene representation learning from
unposed multi-view images. We present \method{}, a framework that places a
strict representation bottleneck between dense visual encoding and downstream tasks. 
Given $N$ unposed images, our encoder represents each view with a small,
fixed set of latent tokens that aggregate information from its dense visual
features.
The resulting scene representation is view-scalable: its
capacity grows with the number of observations, while its per-view token
budget remains independent of image resolution.
These tokens are jointly contextualized across views to form
a compact, scene-dependent representation. Crucially, after this compression
stage, downstream prediction no longer accesses the original dense patch
features. 

Since the contents of a latent scene representation are not specified by a direct
supervision target, we instead learn the representation by decoding it into an
explicit 3D Gaussian scene and supervising the rendered novel views. For this
objective to require the latent memory to carry the scene information needed
for reconstruction, the Gaussian decoder must receive no other input-dependent visual scene content
derived from the input. It therefore has no access to dense patch features, and its Gaussian queries are learned globally across scenes rather than initialized from image features or 3D anchors. The controlled
interface forces all scene-specific visual evidence through the latent memory,
while rendering supervision trains the memory to encode the information
required for novel-view reconstruction.

We build the encoder on a pretrained multi-view geometry foundation
model~\cite{wang2026vggtomega}, which maintains a small number of tokens
per view that participate in cross-view attention alongside its dense
features.  We adopt these tokens as the compact scene representation,
initialize the encoder from the pretrained weights, and optimize it jointly
with the Gaussian decoder. Joint optimization adapts the representation to
the reconstruction objective.

We evaluate the novel-view synthesis quality supported by the learned
representation across different numbers of context views and overlap
regimes on \retenk{}~\cite{zhou2018stereo}. Across the evaluated settings, the compact
representation supports competitive novel-view synthesis.
Experiments across view counts show that the representation incorporates
additional observations through a small, resolution-independent set of
tokens per view.
The \retenk{}~\cite{zhou2018stereo}-trained model also supports
novel-view synthesis on ACID~\cite{liu2021infinite} without fine-tuning,
providing evidence of zero-shot cross-dataset transfer. As a downstream
proof of concept, we train a flow-matching model in the learned token space
and decode its generated latents into explicit 3D Gaussian scenes using the
frozen Gaussian decoder. Together, these results show that
rendering-supervised 3D Gaussian reconstruction is an effective objective
for compact scene representation learning and that the resulting latent
space provides a basis for broader downstream tasks in 3D vision.
\vspace{10pt}
\section{Related Work}
\label{sec:related}
\paragraph{Latent Scene Rrepresentations.}
Scene representations can be learned from multi-view observations
through rendering supervision.~\cite{eslami2018neural} aggregates posed observations into a
fixed-size global scene representation and decodes it directly into the image at a
queried viewpoint.~\cite{kosiorek2021nerf} similarly infers a
global scene latent from posed observations, but decodes it through a
conditional radiance field. Some
approaches~\cite{sajjadi2022srt,sajjadi2023rust,
szymanowicz2026lagernvs,kim2026svsm} retain spatial features for each input view, so their
memory grows with both the number of views and the encoder feature-map
resolution. The encoder--decoder variant of
LVSM~\cite{jin2024lvsm} instead compresses all observations into a fixed set of latent
tokens, and RayZer~\cite{jiang2025rayzer} extends this fixed-latent design to unposed
images. In such designs, additional observations must share the same latent budget.~\cite{wang2026clift} condenses dense features
into ray-anchored tokens and allows the token budget to be selected at
inference. Although these approaches allocate
representational capacity differently, their decoders synthesize each
requested view from target-camera or ray queries, rather than first
decoding an explicit 3D scene.

Other work develops compact 3D token spaces for modeling and
generation. At the object level,~\cite{zhang2023vecset,zhao2023michelangelo,
zhang2024clay,lai2025hunyuan3d25highfidelity3d} learn latent representations of shape
geometry, while appearance is
generated separately from the shape
latent. At the scene level, Nova3R~\cite{chen2026nova3r} and Surflo~\cite{guedon2026surflo} decode fixed global tokens into
geometry, whereas Can3Tok~\cite{gao2025can3tok}
autoencodes preconstructed per-scene 3D
Gaussians. 3DRAE~\cite{wei20263drae} and SceneTok~\cite{asim2026scenetok} learn fixed-size
scene token sets from multi-view observations and model their
distributions in token space. Their target-camera-conditioned decoders
produce RGB images, or RGB images together with point
maps. Other scene generation
systems retain spatially structured latents:~\cite{ren2024scube} uses sparse voxels,
whereas~\cite{yang2025prometheus, huang2026gen3r} uses multi-view RGB-D latents tied to image
grid.

Scene features can also serve as token interfaces for downstream
models.~\cite{hong20233dllm,chen2024ll3da,lin2026qwen3d,
chen2025pts3dllm} provide tokenized scene
features to language models for 3D
understanding. Against this background, our formulation learns a
view-scalable, unstructured scene memory from unposed images through 3D
Gaussian reconstruction, with each input view contributing a small,
resolution-independent token set. A separate Gaussian decoder maps this
memory once into an explicit, renderable 3D scene that carries both
geometry and appearance.
\paragraph{Feed-forward 3D Gaussian Reconstruction.}
Feed-forward 3D Gaussian reconstruction methods predict 3D Gaussian primitives from multi-view
images in a single pass~\cite{zhang2507advances} rather than optimizing a separate 3DGS
representation for each scene~\cite{kerbl2023gaussians}.
A major family ties primitive prediction to the input image 
grid~\cite{szymanowicz2024splatter, tang2024lgm, xu2024grm, zhang2024gs,
szymanowicz2025flash3d},
aggregating cross-view features through epipolar
attention~\cite{charatan2024pixelsplat} or multi-view cost
volumes~\cite{chen2024mvsplat, chen2024mvsplat360, liu2024mvsgaussian,xu2024depthsplat}.
VolSplat~\cite{wang2025volsplat} instead produces Gaussians from a learned 3D voxel grid,
providing a voxel-aligned alternative to image-grid-based
prediction.
Other variants modify the predicted primitive or the inference
procedure through variational feature Gaussians, oriented triangle
primitives, or recurrent
refinement~\cite{wewer2024latentsplat,wang2026trisplat,xu2025resplat}.
Pose-free methods extend feed-forward reconstruction to images without
using camera poses as input~\cite{ye2024noposplat,jiang2025anysplat,
ye2025yonosplat,zhang2025flare,xu2025freesplatter,huang2025no}.
Several build on pretrained geometry models and inherit priors for
multi-view correspondence, camera estimation, and 3D
structure~\cite{wang2024dust3r,leroy2024mast3r,cabon2025must3r,wang2025vggt,
wang2026pi,keetha2026mapanything,burzio2026d,chen2026geometric}.
For pixel-aligned methods, the native primitive count grows with
both the number and resolution of the input views. Subsequent work
reduces this redundancy through fusion, pooling, selection, or
compression~\cite{wang2024freesplat,zhang2024ggn,
song2026tinysplat,park2025ecosplat}.

A second group decouples Gaussian allocation from the input image grid.
Some methods use free learnable
queries~\cite{ren2026tokengs,an2025c3g,li2026querysplat,
itkin2026globalsplat}, while others
derive tokens by clustering visual features or construct them from
depth-based 3D anchors~\cite{veicht2026zipsplat,in2026atsplat,
zhang2026anchorsplat}.
These designs reduce or adapt the Gaussian output budget, but several
retain dense per-view features in the Gaussian-prediction path.
A compact Gaussian output therefore does not by itself
establish a compact latent scene representation. Our method instead
defines a compact, view-scalable scene memory at the encoder--decoder
interface, with a small, resolution-independent token allocation per
view, while its Gaussian output budget is controlled separately.
Rendering-supervised 3D Gaussian reconstruction provides the learning
objective for this memory, while its separation from the Gaussian
output makes it available as a latent interface for downstream 3D
tasks.
\section{Method}
\label{sec:method}
\begin{figure*}[t]
  \vspace{6pt}
  \centering
  \includegraphics[width=1.0\textwidth]{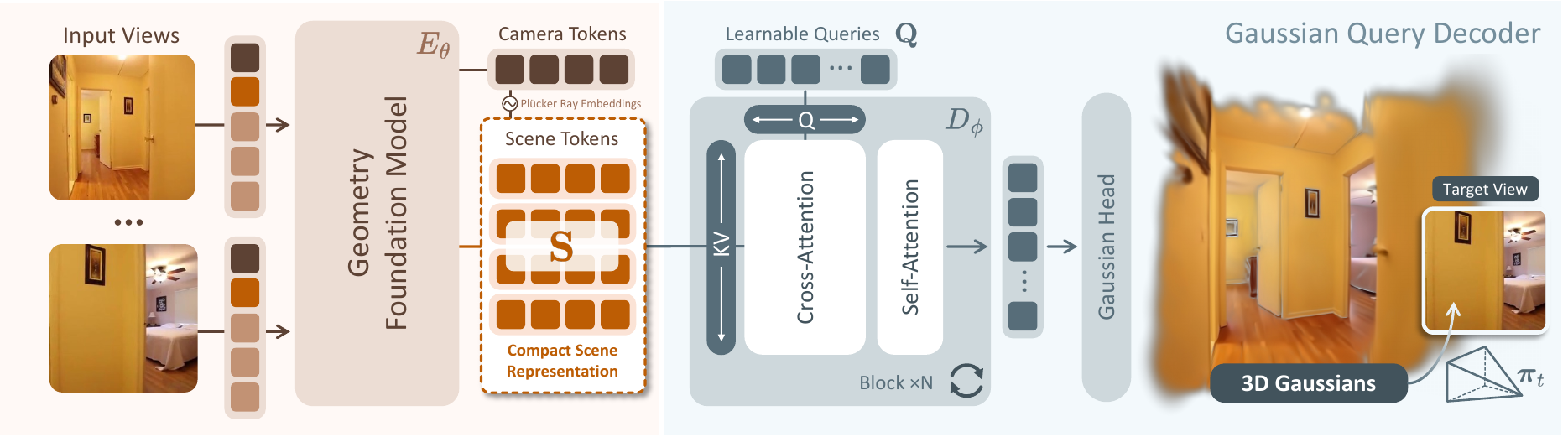}
  \caption{\textbf{Method overview.}
  Our method learns a compact latent representation through
  rendering-supervised 3D Gaussian reconstruction.  The encoder
  $E_{\theta}$, initialized from a pretrained geometry foundation model,
  encodes the unposed context images into one camera token and $n$ scene
  tokens per view.  The camera information anchors the scene tokens to
  form the compact scene representation $\mathbf S$.  Learned
  scene-independent queries $\mathbf Q$ attend to this memory through
  stacked decoder blocks of cross- and self-attention, and a Gaussian
  head maps the final query features to an explicit 3D Gaussian scene.}
  \label{fig:method_overview}
\end{figure*}

Our method learns a compact latent representation that summarizes
scene information across multiple unposed images. We learn this
representation through pose-free 3D Gaussian reconstruction, using novel-view
rendering supervision to optimize it end to end.
\Cref{fig:method_overview} summarizes the pipeline.

Our model first encodes the input views, 
into a
small set of scene tokens, which form the only source of visual scene
content available to the Gaussian decoder.  Learned queries shared across
scenes then retrieve information from this memory and decode it into an
explicit set of 3D Gaussians, which can be rendered from novel viewpoints.
During training, the loss between rendered novel views and their corresponding 
ground-truth target views, together with the auxiliary terms in \cref{sec:method_objective},
is backpropagated through the decoder into the encoder and optimizes
the scene representation itself.

In the following, \Cref{sec:method_setup} formalizes the task setting.
\Cref{sec:method_representation} describes how the encoder constructs
the compact scene representation. \Cref{sec:method_decoder}
presents the Gaussian query decoder, and \cref{sec:method_objective} the
training objective.

\vspace{4pt}
\subsection{Problem Setup}
\label{sec:method_setup}
Given $N$ unposed context images
$\mathcal{I}=\{I_i\}_{i=1}^{N}$, our encoder produces a compact scene memory, from which a scene-agnostic Gaussian decoder $\mathcal{D_{\phi}}$ predicts a set of 3D Gaussians $\mathcal{G}$ by conditioning a set of shared, learnable queries $\mathbf{Q}\in\mathbb{R}^{M\times d}$ on this memory. This decoupling separates the scene-specific representation from the decoding mechanism:
\begin{equation}
    \mathbf{S} = E_{\theta}(\mathcal{I}),
    \quad
    \mathcal{G} = D_{\phi}(\mathbf{Q};\mathbf{S}),
    \quad
    \widehat{I}_{t}
    = \mathcal{R}(\mathcal{G},\boldsymbol{\pi}_{t}),
    \label{eq:overview}
\end{equation}
where $E_{\theta}$ denotes the multi-view encoder and $\mathbf{S}$ represents the latent scene tokens. The decoder $D_{\phi}$ maps the queries $\mathbf{Q}$ into the 3D Gaussian attributes $\mathcal{G}$ conditioned on $\mathbf{S}$. Finally, a differentiable Gaussian splatting renderer $\mathcal{R}$ synthesizes the target view $\widehat{I}_{t}$ given the camera pose $\boldsymbol{\pi}_{t}=(\mathbf{K}_t,\mathbf{T}_t)$, where $\mathbf{K}_t$ and $\mathbf{T}_t$ denote the camera intrinsics and extrinsics, respectively.

\vspace{6pt}
\subsection{Compact Scene Representation}
\label{sec:method_representation}
We build a compact scene representation by coupling a set of view-wise latent scene tokens with corresponding camera tokens.
For view $i$, let $\mathbf z_{i,r}\in\mathbb{R}^{d}$ denote the $r$-th latent scene feature and $\mathbf z_i^{\mathrm{cam}}\in\mathbb{R}^{d}$ the
corresponding camera token, where $d$ is the feature dimension. 
The view-level representation $\mathbf S_i \in \mathbb{R}^{n\times d}$ 
and the full scene representation $\mathbf S \in \mathbb{R}^{[N \cdot n]\times d}$ are constructed as follows:

\begin{equation}
    \mathbf S_i = [\widetilde{\mathbf z}_{i,1},\ldots,\widetilde{\mathbf z}_{i,n}],
    \quad \mathbf{S}=[\mathbf S_1;\ldots;\mathbf S_N].
    \label{eq:scene_representation}
\end{equation}
Here, $\widetilde{\mathbf z}_{i,r}$ represents the camera-conditioned feature derived from $\mathbf z_{i,r}$ and $\mathbf z_i^{\mathrm{cam}}$. The sequence $\mathbf{S}$ aggregates multi-view scene context and acts as the input to the downstream Gaussian decoder.
Each additional view contributes $n$ tokens to $\mathbf S$.
The available representational capacity therefore grows with the number
of observations, while the token allocation per view remains independent
of image resolution.
\paragraph{Backbone Instantiation.}
We instantiate $\mathbf{z}_{i,r}$ and $\mathbf{z}_i^{\mathrm{cam}}$ using the final-layer register and camera tokens from \vggto{} \cite{wang2026vggtomega}. 
For each view, the encoder prepends $n$ learnable register tokens and one camera token to the dense grid of image patch tokens. 
Intra-view frame-attention updates all tokens, whereas inter-frame cross-attention is restricted to camera and register tokens, allowing dense patch tokens to bypass costly cross-view interaction. 
Although standard \vggto{} heads rely on dense patch grids for downstream
predictions, our Gaussian decoder reconstructs the scene from the compact
representation rather than the dense patch grid. We jointly optimize the
backbone and decoder under rendering supervision, so the latent tokens are
learned for reconstruction.

\paragraph{Camera Reference Conditioning.}
Because the extracted latent tokens $\mathbf{z}_{i,r}$ and shared Gaussian queries $\mathbf{Q}$ are not anchored to pixel grids, camera rays, or explicit 3D positions, we explicitly inject a coordinate reference derived from the corresponding camera into each feature. 
Specifically, we project the camera parameters of view $i$ into a Pl\"ucker ray map. 
During training, this map is derived from ground-truth poses $(\mathbf{K}_i, \mathbf{T}_i)$; at inference, it uses predicted parameters $(\widehat{\mathbf{K}}_i, \widehat{\mathbf{T}}_i)$ output by the auxiliary camera head. 
All extrinsics are expressed in a canonical frame anchored at the first
context camera. We normalize the scene scale using the camera-distance
rule described in \cref{sec:scene_normalization}.

We patch-embed and pool the Pl\"ucker ray map into $n$ ray embeddings $\mathbf{e}_{i,r}^{\mathrm{ray}} \in \mathbb{R}^{d}$. The camera-conditioned scene token is then formed via additive fusion:
\begin{equation}
    \widetilde{\mathbf{z}}_{i,r} = \mathbf{z}_{i,r} + \mathbf{e}_{i,r}^{\mathrm{ray}}.
    \label{eq:scene_token_construction}
\end{equation}
To enable unposed inference, the \vggto{} camera head predicts $(\widehat{\mathbf{K}}_i, \widehat{\mathbf{T}}_i)$ from $\mathbf{z}_i^{\mathrm{cam}}$ and the final-layer register tokens $\{\mathbf{z}_{i,r}\}_{r=1}^{n}$. 
During training, we pass stop-gradient features $\{\operatorname{sg}(\mathbf{z}_{i,r})\}_{r=1}^{n}$ to this head, isolating the register features from backpropagation through the auxiliary camera loss. 
As a result, our model operates in an unposed setting during testing without requiring external context poses. Ablations on this conditioning strategy are provided in \cref{sec:exp_conditioning}.
\vspace{2pt}
\subsection{Gaussian Query Decoder}
\label{sec:method_decoder}

The Gaussian decoder maps the compact scene memory $\mathbf{S}$ to a set of 3D Gaussians $\mathcal{G}$ via cross-attention. 
Specifically, the decoder consists of a stack of Transformer layers initialized with $M$ scene-agnostic, learnable queries $\mathbf{Q} \in \mathbb{R}^{M \times d}$. 
To improve computational efficiency, the scene memory $\mathbf{S}$ is projected into key and value representations once and reused across all decoder layers. 
Each layer performs cross-attention over $\mathbf{S}$ and query
self-attention, followed by a feed-forward network. 
Importantly, because the queries $\mathbf{Q}$ are shared across scenes without any input-derived spatial initialization or 3D anchors, all scene-specific information is forcibly bottlenecked through the compact memory $\mathbf{S}$.

A lightweight prediction head subsequently decodes each final query token into $K_{\mathrm{G}}$ 3D Gaussians. 
Each Gaussian is parameterized by its 3D position, appearance coefficients, anisotropic scale, rotation quaternion, and opacity. 
The total number of generated Gaussians is fixed at $N_{\mathrm{G}} = M \cdot K_{\mathrm{G}}$, ensuring a constant computational budget for downstream rendering regardless of the number of context views $N$. 
Further architectural details, parameter settings ($M$, $K_{\mathrm{G}}$, depth), and output parameterization heads are provided in \cref{sec:exp_setup} and the supplementary material.

\vspace{2pt}
\subsection{Training Objective}
\label{sec:method_objective}

Our model is trained end-to-end under joint rendering supervision, geometric regularization, and auxiliary pose supervision. The total loss function is formulated as follows:
\begin{equation}
    \mathcal{L} =
    \mathcal{L}_{\mathrm{rgb}}
    + \lambda_{\mathrm{ssim}}\mathcal{L}_{\mathrm{ssim}}
    + \lambda_{\mathrm{vis}}\mathcal{L}_{\mathrm{vis}}
    + \lambda_{\mathrm{cam}}\mathcal{L}_{\mathrm{cam}},
    \label{eq:loss}
\end{equation}
where $\mathcal{L}_{\mathrm{rgb}}$ denotes the $L_2$ photometric loss between the rendered and ground-truth target views, and $\mathcal{L}_{\mathrm{ssim}}$ measures structural similarity. 

\paragraph{Auxiliary Regularization.}
To eliminate unconstrained Gaussians in invisible regions, we incorporate the visibility regularization loss $\mathcal{L}_{\mathrm{vis}}$ from TokenGS~\cite{ren2026tokengs}, which penalizes Gaussians projecting outside the visual frustums of all supervision cameras. 
Additionally, following \vggto{}~\cite{wang2026vggtomega},
$\mathcal{L}_{\mathrm{cam}}$ is a Smooth $L_1$ loss between the predicted
and ground-truth 9D camera representations. 
We empirically set the loss weights to $\lambda_{\mathrm{ssim}} = 0.2$, $\lambda_{\mathrm{vis}} = 1.0$, and $\lambda_{\mathrm{cam}} = 0.05$.
\vspace{2pt}
\section{Experiments}
\label{sec:experiments}
We first describe the implementation, architecture, and training setup
of \method{} in \cref{sec:exp_setup}. We then evaluate the novel-view
synthesis quality supported by the learned compact scene representation
and examine its zero-shot transfer to
out-of-domain data in \cref{sec:exp_nvs}. Next,
\cref{sec:exp_compactness} quantifies the compactness of the
representation, and \cref{sec:exp_gen} demonstrates its downstream
utility for 3D Gaussian scene generation. Finally,
\cref{sec:exp_ablations} studies the role of camera references in
anchoring the latent scene features and the choice of scene
normalization.

\vspace{6pt}
\subsection{Implementation Details}
\label{sec:exp_setup}
\vspace{2pt}
\subsubsection{Architecture}
We initialize the encoder from the pretrained \vggto{}~\cite{wang2026vggtomega}. 
For each input view, the 16 final-layer register tokens are augmented with pooled Pl\"ucker-ray 
embeddings and constitute the compact scene memory. 
We retain the \vggto{} camera head to estimate the intrinsics and extrinsics of the context views, 
while discarding all dense prediction heads. 
The Gaussian decoder consists of 12 Transformer blocks, each comprising cross-attention 
to the compact scene memory, query self-attention, and an MLP. 
It has a hidden dimension of 1,024 and 16 attention heads, 
and operates on $M=1{,}024$ learned scene-independent queries. 
Each query predicts $K_G=32$ Gaussians, resulting in a fixed output 
budget of $N_G=M K_G=32{,}768$ Gaussians. %
The encoder and Gaussian decoder are optimized jointly end to end.

\vspace{2pt}
\subsubsection{Training}
We train our model on the official RealEstate10K training split~\cite{zhou2018stereo}
using four NVIDIA H100 GPUs with a per-GPU batch size of 6,
corresponding to a total batch size of 24. 
Each training sample is constructed from a temporal window of a training video. 
The first and last frames serve as the context endpoints, 
additional context views are sampled from the intermediate frames, 
and six of the remaining frames are used as rendering targets. 
Training is performed at a resolution of $256\times256$ with a patch size of 16. 
We express all camera poses and scene coordinates in a canonical frame anchored 
at the first context camera and apply camera-distance normalization. 
We use AdamW with $\beta_1=0.9$ and $\beta_2=0.95$, 
together with a OneCycle learning-rate schedule. 

\vspace{6pt}
\subsection{Novel View Synthesis}
\label{sec:exp_nvs}
\vspace{2pt}
\subsubsection{Setup}
We evaluate in-domain
performance on the RealEstate10K test split and assess zero-shot
cross-dataset generalization on ACID \cite{liu2021infinite}, without any finetuning on ACID.
Unless otherwise stated, rendered views are aligned to the ground-truth
target views before the metrics are computed, following
NoPoSplat~\cite{ye2024noposplat}. For a fair comparison, all methods
resize and central-crop each source image to the input resolution used during their training. Target images
and rendered outputs are evaluated at $256\times256$.
The per-method preprocessing and the shared alignment procedure 
are detailed in the supplementary material.

 \begin{table}[t]
  \centering
  \caption{\textbf{Novel view synthesis comparison under various input settings.}
  We report results on \retenk{} with 6--36 context views. \method{}
  achieves competitive \psnr{} and \ssim{} while decoding from a compact
  scene representation, whereas the reconstruction-oriented ZipSplat
  reads dense per-view features and an additional RGB-derived appearance
  path, and attains lower \lpips{}.
  Cells report \psnr{}~$\uparrow$ / \ssim{}~$\uparrow$ /
  \lpips{}~$\downarrow$.
  The \fst{best} and \snd{second-best} results for each metric in every row
  are highlighted.
  }
  \label{tab:coverage_regime}
  \scriptsize
  \setlength{\tabcolsep}{3pt}
  \begin{tabular}{llccc}
    \toprule
    Overlap & $N$ & \textbf{\method{}(ours)} & C3G & ZipSplat \\
    \midrule
    \multirow{4}{*}{small}
      &   6 & \snd{25.91} / \snd{0.838} / 0.245 & 22.81 / 0.734 / \snd{0.240} & \fst{27.81} / \fst{0.870} / \fst{0.142} \\
      & 12 & \snd{27.07} / \snd{0.867} / \snd{0.211} & 23.15 / 0.744 / 0.231 & \fst{28.03} / \fst{0.874} / \fst{0.140} \\
      & 24 & \snd{27.29} / \snd{0.872} / \snd{0.204} & 23.28 / 0.749 / 0.229 & \fst{28.09} / \fst{0.875} / \fst{0.138} \\
      & 36 & \snd{27.26} / \snd{0.871} / \snd{0.205} & 23.28 / 0.749 / 0.229 & \fst{28.09} / \fst{0.876} / \fst{0.137} \\
    \midrule
    \multirow{4}{*}{medium}
      &   6 & \snd{26.60} / \snd{0.850} / 0.236 & 23.36 / 0.737 / \snd{0.223} & \fst{28.30} / \fst{0.873} / \fst{0.130} \\
      & 12 & \snd{27.64} / \snd{0.875} / \snd{0.207} & 23.67 / 0.746 / 0.217 & \fst{28.44} / \fst{0.876} / \fst{0.129} \\
      & 24 & \snd{27.83} / \fst{0.879} / \snd{0.201} & 23.79 / 0.749 / 0.215 & \fst{28.50} / \snd{0.877} / \fst{0.127} \\
      & 36 & \snd{27.80} / \fst{0.879} / \snd{0.202} & 23.79 / 0.749 / 0.215 & \fst{28.48} / \snd{0.877} / \fst{0.127} \\
    \midrule
    \multirow{4}{*}{large}
      &   6 & \snd{27.10} / \snd{0.852} / 0.238 & 23.42 / 0.714 / \snd{0.225} & \fst{28.22} / \fst{0.858} / \fst{0.130} \\
      & 12 & \snd{27.95} / \fst{0.872} / \snd{0.215} & 23.65 / 0.721 / 0.220 & \fst{28.32} / \snd{0.860} / \fst{0.128} \\
      & 24 & \snd{28.06} / \fst{0.875} / \snd{0.211} & 23.71 / 0.723 / 0.219 & \fst{28.32} / \snd{0.860} / \fst{0.128} \\
      & 36 & \snd{28.04} / \fst{0.875} / \snd{0.211} & 23.69 / 0.722 / 0.220 & \fst{28.32} / \snd{0.860} / \fst{0.128} \\
    \bottomrule
  \end{tabular}
  \vspace{10pt}
\end{table}

\begin{figure*}[t]
  \vspace{8pt}
  \centering
  \setlength{\tabcolsep}{0pt}
  \renewcommand{\arraystretch}{0}

  \let\originalqualhimage\qualhimage
  \renewcommand{\qualhimage}[1]{%
    \adjustbox{valign=m}{\originalqualhimage{#1}}%
  }

  \renewcommand{\qualhlabel}[1]{%
    \adjustbox{valign=m}{%
      \rotatebox[origin=c]{90}{ \footnotesize \strut #1}%
    }%
  }
  \resizebox{\linewidth}{!}{%
  \begin{tabular}{@{}c@{\hspace{1.4mm}}c@{\hspace{0.5mm}}c@{\hspace{0.5mm}}c@{\hspace{2.2mm}}c@{\hspace{0.5mm}}c@{\hspace{0.5mm}}c@{\hspace{2.2mm}}c@{\hspace{0.5mm}}c@{\hspace{0.5mm}}c@{\hspace{2.2mm}}c@{\hspace{0.5mm}}c@{\hspace{0.5mm}}c@{}}
    & \multicolumn{3}{c@{\hspace{2.2mm}}}{\qualhgroup{6 views}}
    & \multicolumn{3}{c@{\hspace{2.2mm}}}{\qualhgroup{12 views}}
    & \multicolumn{3}{c@{\hspace{2.2mm}}}{\qualhgroup{24 views}}
    & \multicolumn{3}{c@{}}{\qualhgroup{36 views}} \\[1.2mm]

  \qualhlabel{GT}
  & \qualhimage{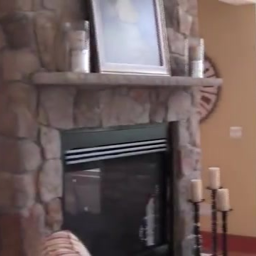}
  & \qualhimage{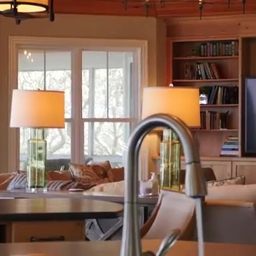}
  & \qualhimage{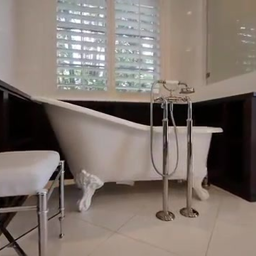}
  & \qualhimage{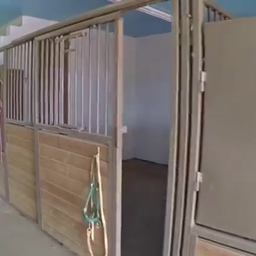}
  & \qualhimage{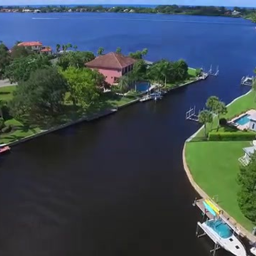}
  & \qualhimage{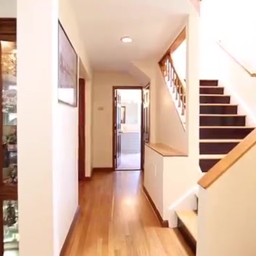}
  & \qualhimage{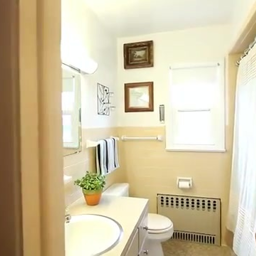}
  & \qualhimage{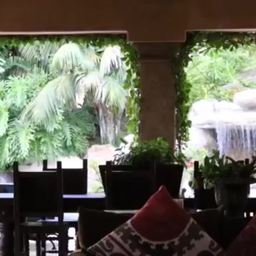}
  & \qualhimage{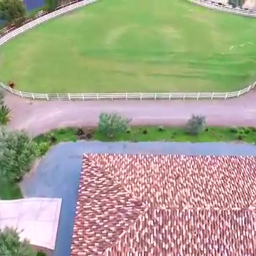}
  & \qualhimage{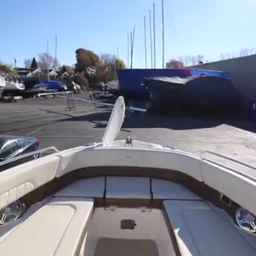}
  & \qualhimage{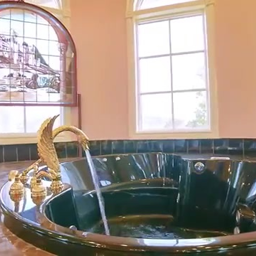}
  & \qualhimage{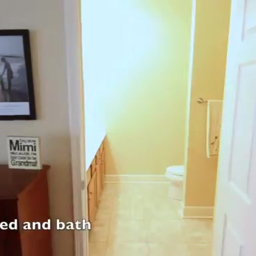} \\[0.7mm]

  \qualhlabel{\bfseries \method{}}
  & \qualhimage{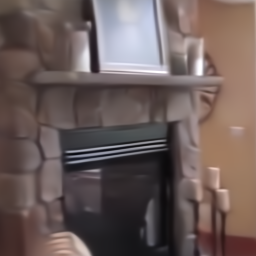}
  & \qualhimage{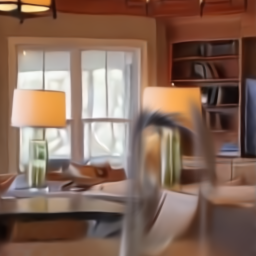}
  & \qualhimage{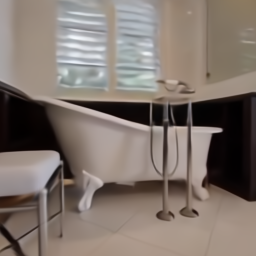}
  & \qualhimage{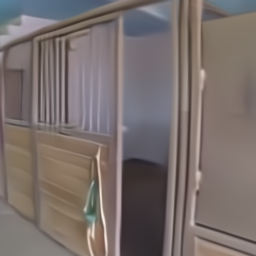}
  & \qualhimage{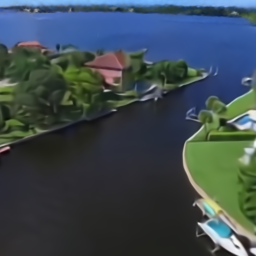}
  & \qualhimage{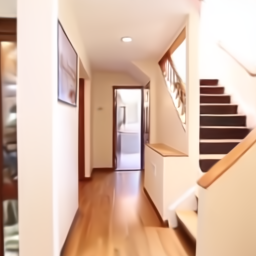}
  & \qualhimage{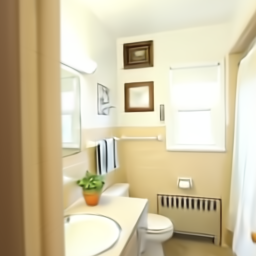}
  & \qualhimage{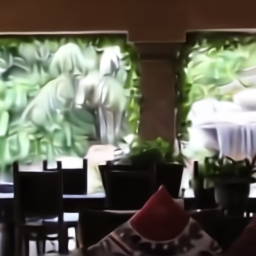}
  & \qualhimage{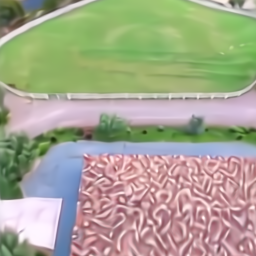}
  & \qualhimage{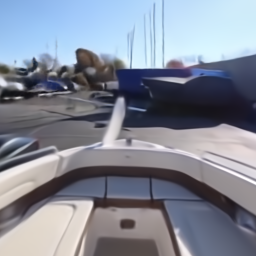}
  & \qualhimage{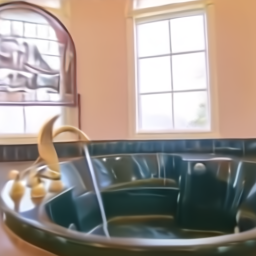}
  & \qualhimage{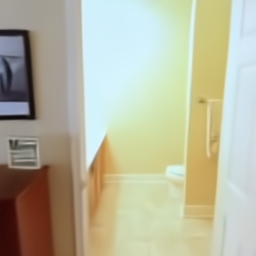} \\[0.7mm]

  \qualhlabel{C3G}
  & \qualhimage{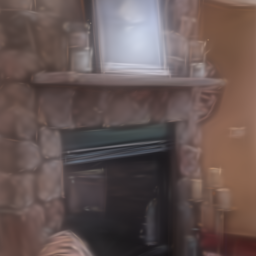}
  & \qualhimage{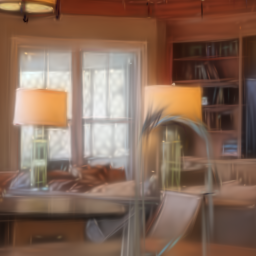}
  & \qualhimage{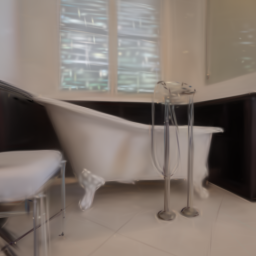}
  & \qualhimage{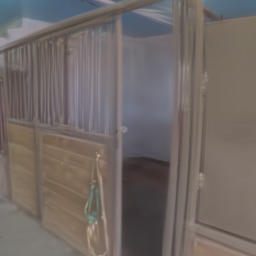}
  & \qualhimage{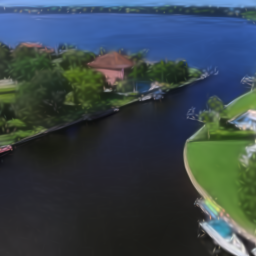}
  & \qualhimage{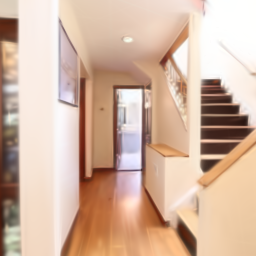}
  & \qualhimage{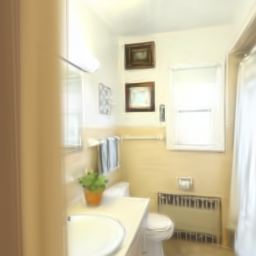}
  & \qualhimage{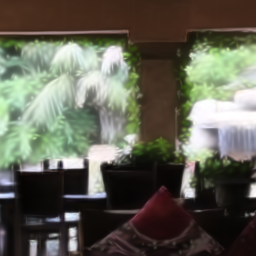}
  & \qualhimage{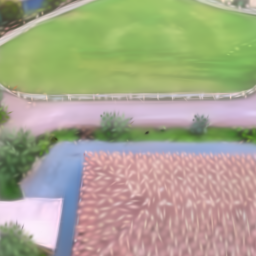}
  & \qualhimage{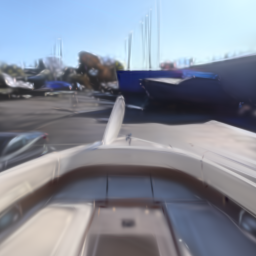}
  & \qualhimage{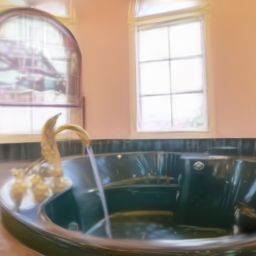}
  & \qualhimage{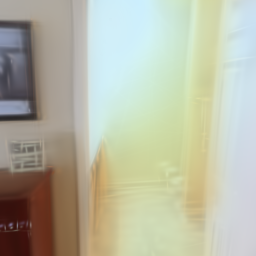} \\[0.7mm]

  \qualhlabel{ZipSplat}
  & \qualhimage{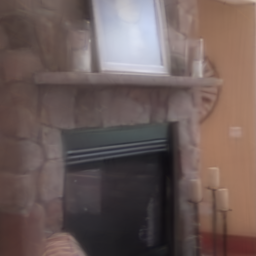}
  & \qualhimage{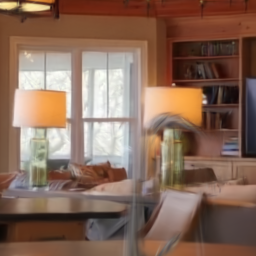}
  & \qualhimage{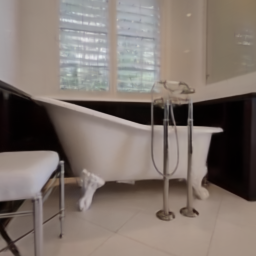}
  & \qualhimage{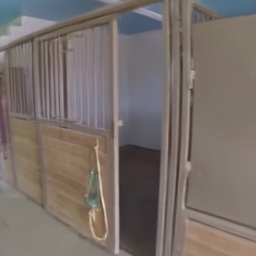}
  & \qualhimage{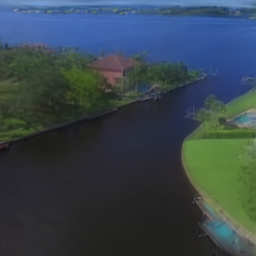}
  & \qualhimage{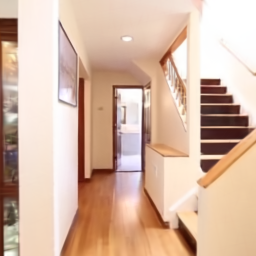}
  & \qualhimage{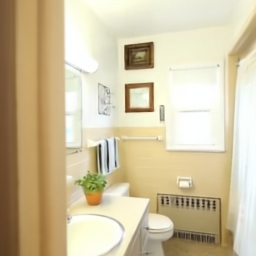}
  & \qualhimage{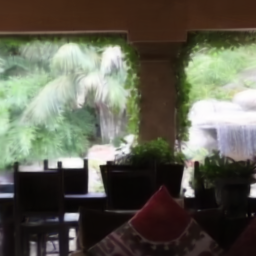}
  & \qualhimage{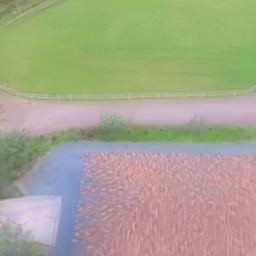}
  & \qualhimage{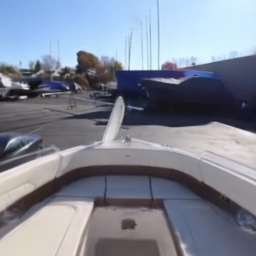}
  & \qualhimage{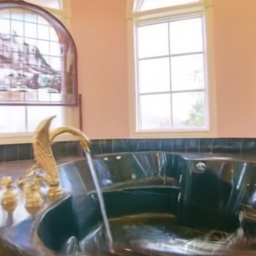}
  & \qualhimage{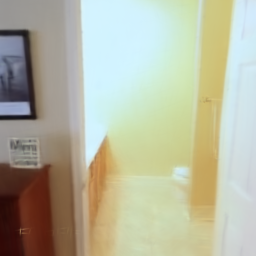}
  \end{tabular}
  }
  \caption{\textbf{Qualitative novel-view synthesis with increasing context-view counts.}
  For each setting, columns show representative target views, while rows show the ground truth and predictions from \method{}, C3G, and ZipSplat. Using only the compact scene tokens at the encoder--decoder interface, \method{} preserves the global scene structure and appearance more faithfully than C3G and approaches the dense-memory ZipSplat baseline as more context views are provided. ZipSplat retains finer local textures in some regions, consistent with the remaining \lpips{} gap in \cref{tab:coverage_regime}.}
  \label{fig:qualitative_h}
\end{figure*} 

\begin{table}[t]
  \centering
  \caption{\textbf{Zero-shot transfer to ACID.}
  Cells report \psnr{}~$\uparrow$ / \ssim{}~$\uparrow$ /
  \lpips{}~$\downarrow$ over the indexed target views.
  The \fst{best} and \snd{second-best} results for each metric in every row
  are highlighted.}
  \label{tab:acid_zeroshot}
  \scriptsize
  \setlength{\tabcolsep}{6pt}
  \begin{tabular}{lccc}
    \toprule
    $N$ & \textbf{\method{}(ours)} & C3G & ZipSplat \\
    \midrule
    12 & \fst{26.12} / \fst{0.771} / 0.366 & 22.25 / 0.598 / \snd{0.331} & \snd{24.90} / \snd{0.718} / \fst{0.244} \\
    24 & \fst{26.24} / \fst{0.776} / 0.361 & 22.24 / 0.599 / \snd{0.331} & \snd{24.90} / \snd{0.718} / \fst{0.244} \\
    36 & \fst{26.22} / \fst{0.776} / 0.361 & 22.21 / 0.598 / \snd{0.332} & \snd{24.86} / \snd{0.717} / \fst{0.244} \\
    \bottomrule
  \end{tabular}
\end{table}

\vspace{4pt}
\subsubsection{Evaluation Protocols and Metrics}
Following the C3G evaluation protocol~\cite{an2025c3g}, we start from
the NoPoSplat~\cite{ye2024noposplat} two-view \retenk{}~\cite{zhou2018stereo} index and
uniformly insert frames between the two indexed context endpoints.
This produces context sets of 6, 12, 24, and 36 views while keeping the
three indexed target views unchanged. Novel-view rendering quality is
evaluated using \psnr{}, \ssim{}, and \lpips{}.

\subsubsection{Baselines}
To contextualize the novel-view rendering 
quality supported by the learned representation We report 
two pose-free methods for generalizable novel view synthesis 
that decode global Gaussian sets as reference points: 
C3G~\cite{an2025c3g} and ZipSplat~\cite{veicht2026zipsplat}. 
C3G uses a fixed output budget of 2K Gaussians. We configure ZipSplat with the same 
32K-Gaussian output budget as \method{}.

\subsubsection{Reconstruction Quality}
As shown in \cref{tab:coverage_regime}, reconstruction quality improves
most from 6 to 12 context views across all three overlap regimes and
changes little beyond 24 views. The endpoint views and Gaussian output
budget are held fixed. Under this control, the additional observations
can affect the Gaussian decoder only through the compact scene
representation. The improvement therefore provides direct evidence that the memory integrates additional visual evidence across views and retains both geometric and appearance information useful for rendering.

C3G provides a lower-output-budget reference with 2K Gaussians;
\method{} achieves higher \psnr{} and \ssim{} across all settings and
lower \lpips{} with at least 12 context views. ZipSplat provides a
dense-memory reference at the same Gaussian output budget. With 12 or
more context views in the large-overlap regime, \method{} approaches
ZipSplat in \psnr{} and exceeds it in \ssim{}, although a clear
\lpips{} gap remains. Qualitative comparisons 
in \cref{fig:qualitative_h} are consistent with
this metric pattern: \method{} retains the overall scene structure and
appearance, while ZipSplat preserves finer local textures in some
regions.

The differences between \method{} and ZipSplat should also be
interpreted in light of their distinct decoder interfaces. ZipSplat is
designed primarily for 3D reconstruction and does not isolate a compact
latent scene representation at the encoder--decoder interface: its
decoder reads dense per-view features, and its Gaussian head
additionally receives color features obtained by cross-attending
directly to patch-embedded RGB inputs. In contrast, we explicitly
separate the compact scene representation from the Gaussian decoder
that reads it. This separation exposes a reusable latent
interface for downstream modeling beyond reconstruction.

The results in \cref{tab:acid_zeroshot} show that the
\retenk{}-trained model supports novel-view synthesis on ACID without
fine-tuning, providing evidence of zero-shot cross-domain transfer.
Together, these results show that rendering-supervised 3D Gaussian
reconstruction learns a compact scene memory that retains sufficient
visual information for pose-free novel-view synthesis.

\begin{figure}[t]
  \centering
  \includegraphics[width=\linewidth]{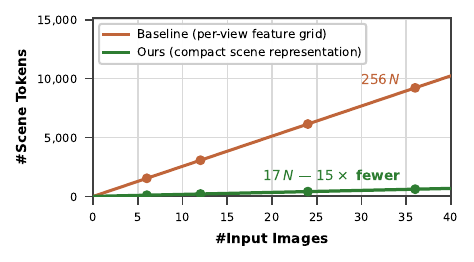}
  \caption{\textbf{Token-count scaling with the number of input views.}
  For methods that use a dense per-view feature grid as the Gaussian
  decoder's keys and values, a $16\times16$ grid yields $256N$ tokens.
  Our compact scene representation uses 17 tokens per view and grows as
  $17N$, approximately $15\times$ fewer.}
  \label{fig:scene_tokens}
\end{figure}
\subsection{Compactness Comparison}
\label{sec:exp_compactness}
Beyond reconstruction quality, we examine the token count of the learned scene 
representation and how it scales with the number of input views.
\Cref{fig:scene_tokens} compares our compact scene representation with
the dense-grid interface used by methods whose Gaussian decoder attends
to full per-view features as keys and values. A $16\times16$ feature
grid contributes 256 key--value tokens per view, so the decoder context
grows as $256N$. In contrast, our representation uses 17 tokens per
view and grows as $17N$, approximately $15\times$ fewer across the
evaluated view counts.

\vspace{6pt}
\subsection{3D Scene Generation}
\label{sec:exp_gen}
\begin{figure*}[t]
\vspace{8pt}
  \centering
  \setlength{\tabcolsep}{0pt}
  \renewcommand{\arraystretch}{0}

  \resizebox{\linewidth}{!}{%
  \begin{tabular}{@{}c@{\hspace{0.8mm}}c@{\hspace{0.8mm}}c@{\hspace{0.8mm}}c@{\hspace{3.5mm}}c@{\hspace{0.8mm}}c@{\hspace{0.8mm}}c@{\hspace{0.8mm}}c@{}}
    \genhead{Condition}
    & \genhead{Generated View 0}
    & \genhead{Generated View 1}
    & \genhead{Generated View 2}
    & \genhead{Condition}
    & \genhead{Generated View 0}
    & \genhead{Generated View 1}
    & \genhead{Generated View 2} \\[1.2mm]

  \genimage{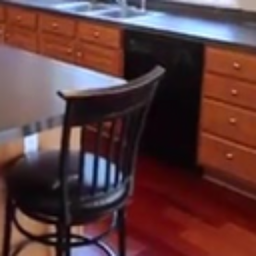}
  & \genimage{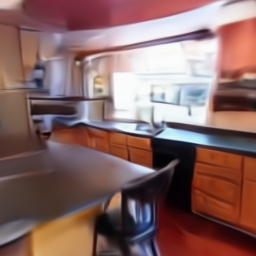}
  & \genimage{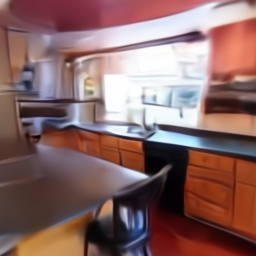}
  & \genimage{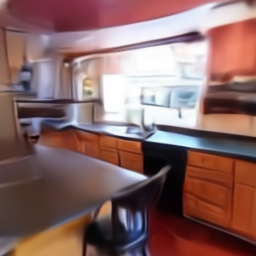}
  & \genimage{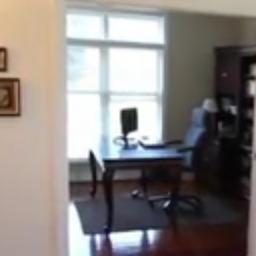}
  & \genimage{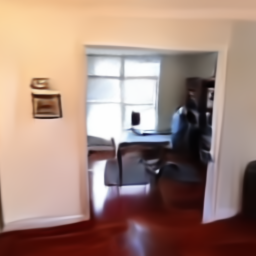}
  & \genimage{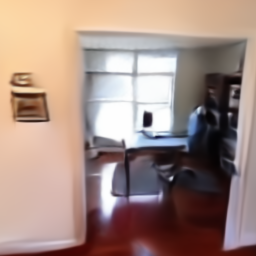}
  & \genimage{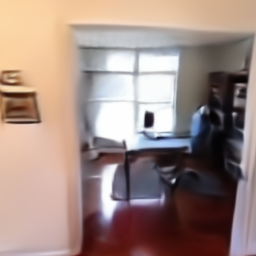} \\[1.0mm]

  \genimage{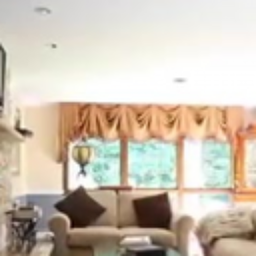}
  & \genimage{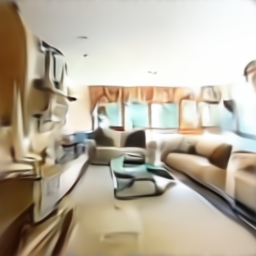}
  & \genimage{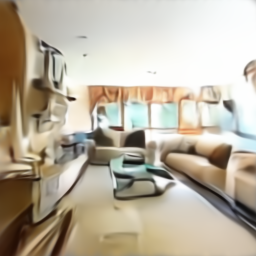}
  & \genimage{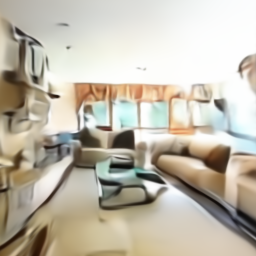}
  & \genimage{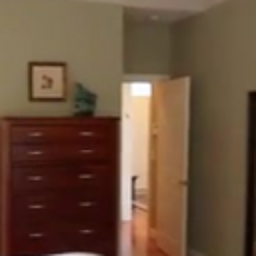}
  & \genimage{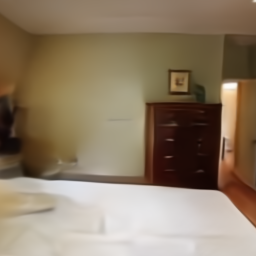}
  & \genimage{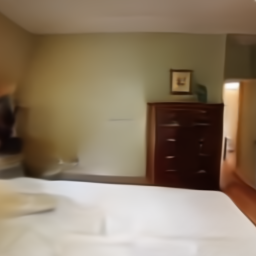}
  & \genimage{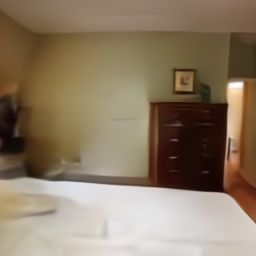} \\[1.0mm]

  \genimage{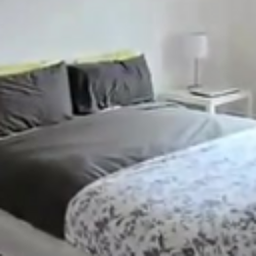}
  & \genimage{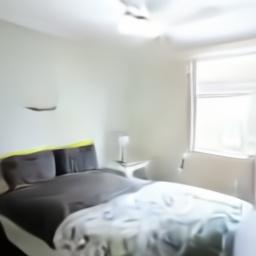}
  & \genimage{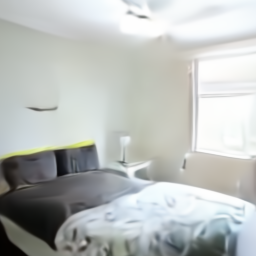}
  & \genimage{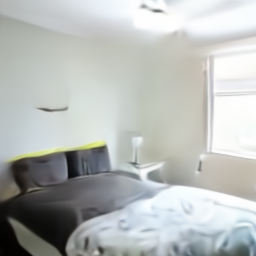}
  & \genimage{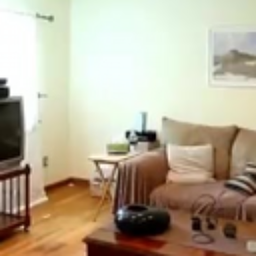}
  & \genimage{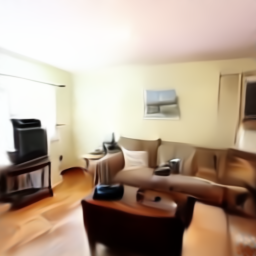}
  & \genimage{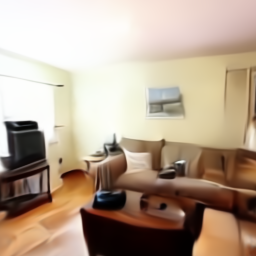}
  & \genimage{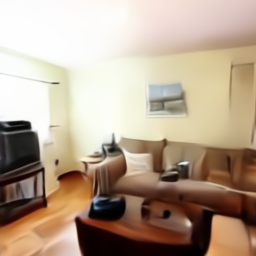}
  \end{tabular}
  }
  \caption{\textbf{Conditional 3D Gaussian scene generation in the
  learned compact latent space.}
  For each of six samples, a four-column group shows one conditioning
  image followed by three novel views.  A flow-matching model generates
  the compact latent, which the frozen Gaussian decoder decodes into an
  explicit 3D Gaussian representation.  The three output images are
  novel views rendered from this representation.}
  \label{fig:generation}
\end{figure*}
 We further demonstrate that the compact, unstructured scene
representation learned through reconstruction can serve as the latent
space of a generative model~\cite{rombach2022high} for 3D scene
generation. We use the trained reconstruction model as a frozen first
stage. During second-stage training, the frozen encoder maps eight views of
each scene to the latent scene features
$\mathbf Z\in\mathbb{R}^{(8n)\times d}$. These features serve as the training targets for a
second-stage flow-matching model~\cite{lipman2023flow}, which generates
$\mathbf Z$ conditioned on the first of the eight images supplied to
the encoder, together with the camera parameters of all eight views.
The network architecture, conditioning mechanism, and training details of the flow-matching model are provided in the supplementary material.

At inference, the generated features are combined with the
corresponding camera references to form the scene representation
$\mathbf S$, following \cref{eq:scene_token_construction}. The frozen
Gaussian decoder then maps $\mathbf S$ to an explicit \gs{} scene,
which can be rendered from novel viewpoints in real time.
\Cref{fig:generation} shows six generated scenes, each rendered from
multiple viewpoints. The samples show plausible scene completions that
remain coherent across the rendered views, providing evidence that the
learned token representation supports downstream 3D scene generation.
\vspace{6pt}
\subsection{Ablation Study}
\label{sec:exp_ablations}
This section examines two design choices for learning the compact,
unstructured scene representation: the camera reference used to anchor
its latent features and scene-scale normalization.
\vspace{6pt}
\subsubsection{Anchoring Latent Scene Features}
\label{sec:exp_conditioning}
We ablate the camera reference used to anchor the latent scene features.
We train three variants while keeping all other settings identical. In
the \emph{scene features only} variant, the decoder reads the latent scene
features without a camera reference. In the \emph{global camera anchor}
variant, each view's camera token is concatenated with its latent scene
features. In our \emph{pooled ray-map references} variant, each latent scene
feature is associated with a pooled ray-map reference derived from its source
camera, as described in \cref{sec:method_representation}.
We compare validation \psnr{} and
the fraction of active Gaussians (opacity above 0.1) at the same
training step.  The first two variants behave almost identically:
relative to the ray embeddings, their \psnr{} drops by about 2.5\,dB, and
the fraction of active Gaussians is more than ten percentage points lower.
Without a coordinate reference, the rendering loss reduces a poorly
placed Gaussian's opacity more readily than it relocates the Gaussian.
As the opacity approaches zero, the Gaussian receives progressively
weaker rendering gradients and remains inactive, so reconstruction stays blurry.
Our pooled ray-map construction provides a coarse coordinate reference from the start of training and mitigates this
failure mode.
\vspace{6pt}
\subsubsection{Scene Normalization}
\label{sec:scene_normalization}
We now compare three difference scene normalization
choices. Our \emph{camera-distance normalization} uses
$\rho_{\mathrm{scene}}=\max_i\lVert\mathbf c_i-\mathbf c_1\rVert_2$
---the maximum distance from any context camera to the first---as the
scene scale. We divide all camera translations and scene coordinates by
this scale. The
\emph{point-cloud-radius normalization} instead uses the 95th-percentile
radius of the point cloud obtained by unprojecting the precomputed
context-view depth maps. The \emph{no normalization} variant retains the raw
dataset scale.
The point-cloud and unnormalized variants lead to less stable early
optimization, with the latter showing the largest degradation.
Camera-distance normalization instead defines a consistent scale from
the context-camera configuration. We therefore use camera-distance normalization in all experiments.
\vspace{6pt}
\section{Conclusion}
\label{sec:conclusion}
In this work, we present \method{}, a framework for learning a compact, view-scalable scene representation directly from unposed multi-view images via 3D Gaussian reconstruction. By compressing multi-view inputs into a concise set of view-wise tokens, \method{} establishes an explicit information bottleneck: the scene memory serves as the sole condition for a set of scene-agnostic queries, forcing it to retain only the essential scene tokens required for high-fidelity geometric and appearance reconstruction.

Furthermore, we demonstrate that this compact token representation provides an effective latent space for 3D generative modeling. Compared to video-diffusion-based scene synthesis, generating scene latents directly in this space significantly accelerates inference while enforcing strict cross-view geometric consistency. Once generated, the latent representation is decoded in a single feed-forward pass into an explicit 3D Gaussian scene, enabling real-time novel-view rendering without the per-view decoding overhead inherent to prior approaches.

Looking forward, we anticipate several promising directions. 
Future work includes scaling up the pre-training data to explore whether this compact representation is able to generalize to open-world environments. 
Additionally, we plan to extend the generative latent space to support flexible view counts and multi-modal controls (\eg, text conditioning). More broadly, we envision this compact scene memory acting as a general-purpose, reusable interface for downstream 3D understanding, interactive editing, and persistent spatial memory in embodied intelligence.

{
    \small
    \bibliographystyle{ieeenat_fullname}
\setlength{\bibsep}{0pt}
    \bibliography{main}

}

\clearpage
\maketitlesupplementary

\appendix

\section{Overview}
\label{sec:supp_overview}
This supplementary material complements the main paper with additional
implementation details, evaluation protocols, and analyses of the compact
scene representation used for reconstruction and generation.
\Cref{sec:supp_impl_eval} describes the Gaussian output parameterization
and training schedule, while \cref{sec:supp_eval_protocol} details the
image preprocessing and pose-alignment procedures used for evaluation.
\Cref{sec:supp_visual} presents additional qualitative
comparisons across context-view counts and overlap regimes.
Finally, \cref{sec:supp_generation} describes the latent construction,
conditioning, training, and sampling procedure used for 3D scene
generation.

\section{Additional Implementation Details}
\label{sec:supp_impl_eval}

\subsection{Gaussian Output Head}
We build on the Gaussian output head of TokenGS~\cite{ren2026tokengs}.
A linear prediction layer maps each decoded query feature to 32 raw
Gaussian parameter vectors, which are mapped through attribute-specific
activation functions.  The raw center outputs use the signed-exponential
activation $\operatorname{sign}(x)(e^{|x|}-1)$.  We then add a constant
offset of one to the predicted $z$ coordinate, placing the initial centers
near one normalized scene unit along the positive optical axis ($+z$) of
the first context camera.  The raw scale outputs are passed through the
shifted, clipped exponential $\min(e^{x-s},c)$, where $c=0.075$ and
$s=1-\log c$.  Each raw opacity logit $x$ is mapped to $\sigma(x-2)$,
and the four raw rotation outputs are normalized to form a unit
quaternion.  For appearance, the three raw color values are mapped to RGB range $(0,1)$ using $\tfrac{1}{2}\tanh(x)+\tfrac{1}{2}$.
The weights of the linear prediction layer are initialized from a truncated
normal distribution with standard deviation $0.002$, and its biases are
initialized to zero.  

\subsection{Training Details}
The complete model comprises the \vggto{}~\cite{wang2026vggtomega} encoder, a 12-layer Gaussian
decoder, and the retained camera head, with $1{,}116$\,M parameters in
total. We initially train the model for $500{,}000$ steps using AdamW
($\beta_1=0.9$, $\beta_2=0.95$) and a OneCycle learning-rate schedule
with a peak learning rate of $4\times10^{-4}$ and a $5{,}000$-step
warm-up. This training uses eight context views and spherical harmonics
of degree zero.
We subsequently fine-tune the model using a randomized number of context views and spherical harmonics of degree three. We initialize the DC coefficients from the learned degree-zero color readout and set the newly introduced higher-order coefficients to zero. For each batch, the context-view count is sampled uniformly from $6$--$12$ with probability $0.75$ or from $13$--$24$ with probability $0.25$.
\section{Evaluation Protocol}
\label{sec:supp_eval_protocol}
\subsection{Image Preprocessing}
We prepare the input images for each method at the resolution used
during its training. Starting from the original $640\times360$ frames,
\method{} and C3G~\cite{an2025c3g} resize each frame to a height of 256 pixels while
preserving its aspect ratio, followed by a $256\times256$ center crop.
ZipSplat~\cite{veicht2026zipsplat} follows the same procedure at its native training resolution,
resizing each original frame to a height of 252 pixels before taking a
$252\times252$ center crop. 

\subsection{Pose Alignment}
Pose alignment and metric evaluation are performed at
$256\times256$ for all methods. For ZipSplat~\cite{veicht2026zipsplat}, target images processed
at its native $252\times252$ resolution are upsampled to
$256\times256$ for evaluation, while its reconstructed scene is
rendered directly at $256\times256$.

The ground-truth target cameras are first transformed into each
method's canonical reconstruction frame according to its scene
canonicalization and normalization convention.
Starting from these transformed target cameras, we refine each target
pose photometrically while keeping the reconstructed Gaussians fixed.
For C3G~\cite{an2025c3g} and ZipSplat~\cite{veicht2026zipsplat} we use their officially released evaluation scripts.
Across all methods, we use shared alignment settings: rotation and
translation learning rates of $0.005$, $100$ optimization steps, and no
early stopping.

\section{Additional Qualitative Results}
\label{sec:supp_visual}
\Cref{fig:supp_visual} provides additional qualitative comparisons at four
context-view counts, over the same overlap regimes.
\newlength{\suppvisw}
\setlength{\suppvisw}{16mm}
\newcommand{\suppvisimg}[1]{\adjustbox{valign=m}{\includegraphics[width=\suppvisw]{#1}}}
\newcommand{\suppvislabel}[1]{\adjustbox{valign=m}{\rotatebox[origin=c]{90}{\small\strut #1}}}
\newcommand{\suppvisgroup}[1]{{\small #1}}
\newcommand{\suppvishead}[1]{{\normalsize\bfseries #1}}

\newcommand{\suppvisrow}[2]{%
      & \suppvisimg{figures/supp_visual/#1v_small1_#2.png}
      & \suppvisimg{figures/supp_visual/#1v_small2_#2.png}
      & \suppvisimg{figures/supp_visual/#1v_small3_#2.png}
      & \suppvisimg{figures/supp_visual/#1v_medium1_#2.png}
      & \suppvisimg{figures/supp_visual/#1v_medium2_#2.png}
      & \suppvisimg{figures/supp_visual/#1v_medium3_#2.png}
      & \suppvisimg{figures/supp_visual/#1v_large1_#2.png}
      & \suppvisimg{figures/supp_visual/#1v_large2_#2.png}
      & \suppvisimg{figures/supp_visual/#1v_large3_#2.png}%
}

\newcommand{\suppvispage}[1]{%
  \centering
  \setlength{\tabcolsep}{0pt}%
  \renewcommand{\arraystretch}{0}%
  \resizebox{\linewidth}{!}{%
  \begin{tabular}{@{}c@{\hspace{1.4mm}}ccc@{\hspace{2.2mm}}ccc@{\hspace{2.2mm}}ccc@{}}
    & \multicolumn{9}{c@{}}{\suppvishead{#1 context views}} \\[1.0mm]
    & \multicolumn{3}{c@{\hspace{2.2mm}}}{\suppvisgroup{Small overlap}}
    & \multicolumn{3}{c@{\hspace{2.2mm}}}{\suppvisgroup{Medium overlap}}
    & \multicolumn{3}{c@{}}{\suppvisgroup{Large overlap}} \\[1.2mm]
    \suppvislabel{GT}              \suppvisrow{#1}{target}   \\[0.6mm]
    \suppvislabel{\textbf{\method{}}} \suppvisrow{#1}{ours}     \\[0.6mm]
    \suppvislabel{C3G}             \suppvisrow{#1}{c3g}      \\[0.6mm]
    \suppvislabel{ZipSplat}        \suppvisrow{#1}{zipsplat} \\
  \end{tabular}}%
}

\begin{figure*}[p]
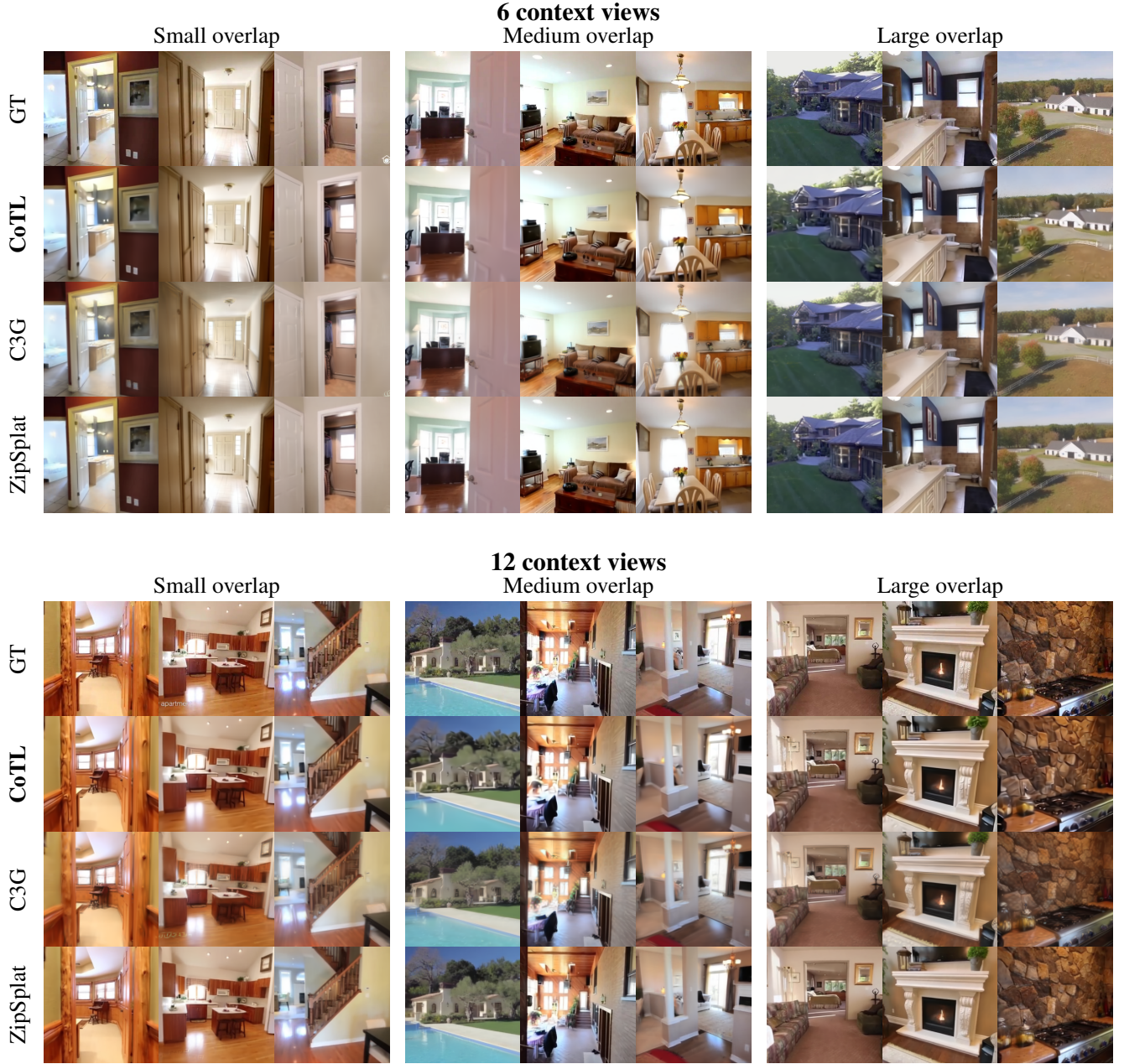

  \suppvispage{6}
  \par\vspace{6mm}
  \suppvispage{12}
  \caption{\textbf{Additional novel-view synthesis results.}
  Each group is one setting of $N$, the number of context views the model
  receives as input, stated above the group.  Within a group the nine columns
  are three scenes from each of three overlap bands
  where overlap is the value measures how much of the scene those views share.
  Rows are the ground truth and the renderings of \method{}, C3G and ZipSplat
  from the same held-out viewpoint.  Continued on the next page with $N = 24$
  and $N = 36$.}
  \label{fig:supp_visual}
\end{figure*}

\begin{figure*}[p]
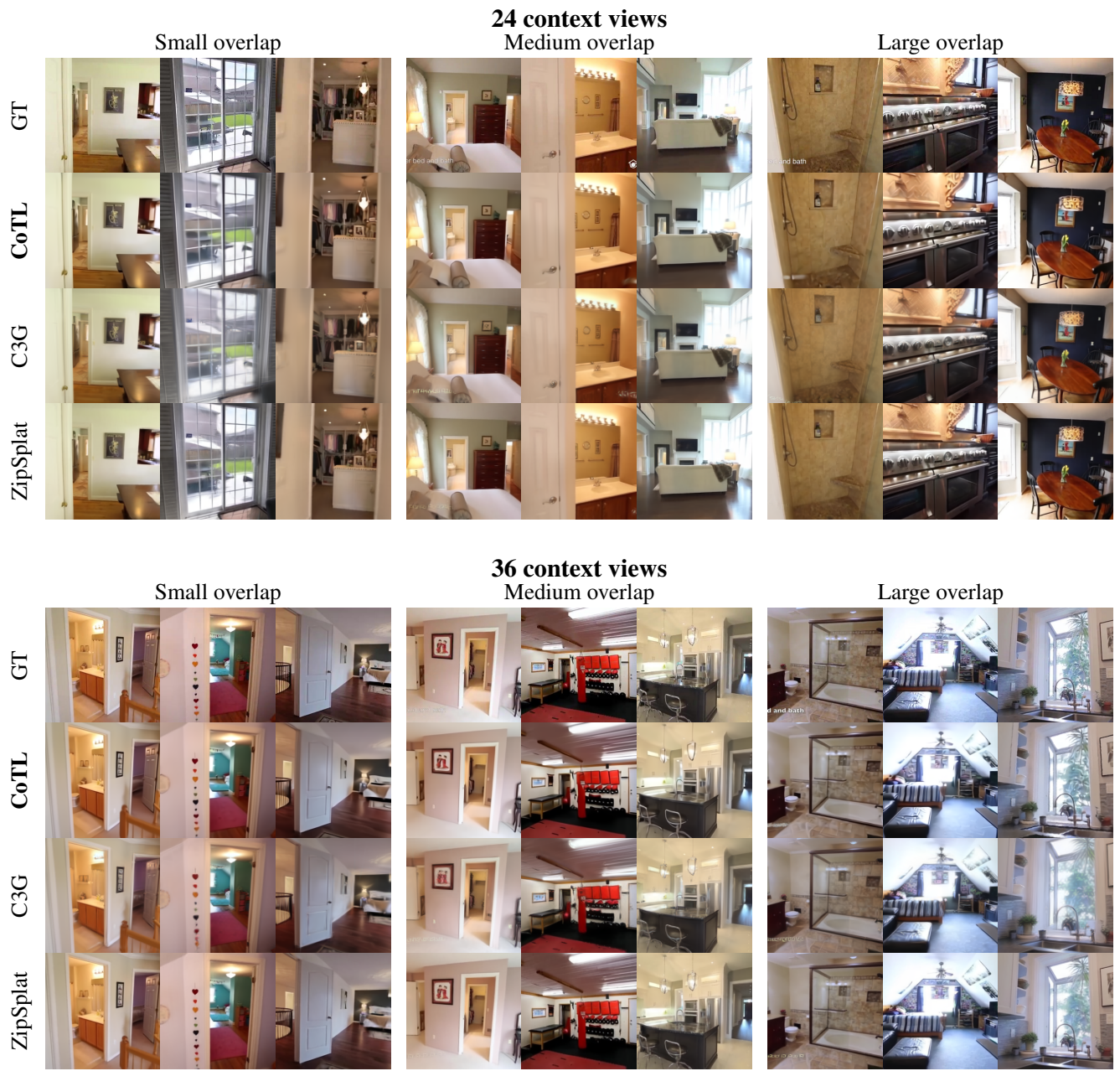

  \ContinuedFloat
  \suppvispage{24}
  \par\vspace{6mm}
  \suppvispage{36}
  \caption[]{\textbf{Additional novel-view synthesis results (continued).}
  Layout as on the previous page.}
\end{figure*}

\section{Scene Generation}
\label{sec:supp_generation}
The generative model performs single-image-conditioned 3D scene
generation. The reference image provides the observed scene content,
while the camera parameters of eight views define the spatial extent of
the generated scene. Conditioned on both, the model produces a 3D scene
that can be rendered from novel viewpoints throughout this extent.

\paragraph{Latent Space.}
The flow-matching model operates on the latent scene features
$\mathbf Z$ of the reconstruction model. The reconstruction encoder maps
eight views of each scene to $8\times16=128$ features of width $1024$,
extracted before the camera reference in
\cref{eq:scene_token_construction} is added. 
We normalize the features per
channel using statistics estimated over the training split and denote
their interpolated state at flow time $t$ by $\mathbf Z_t$. The model is
conditioned on the first of the eight images and the camera parameters of
all eight views, expressed in the canonical frame anchored at the first
camera.

\paragraph{Flow-matching Network.}
The flow-matching network follows the transformer architecture of
LightningDiT~\cite{yao2025lightningdit}. Its backbone contains $12$
blocks of width $1024$ with $16$ attention heads. 
Following~\cite{wei20263drae, zheng2026rae}, 
we append a two-block DDT head \cite{wang2026ddt} of width $2048$ to
 the backbone for high-dimensional velocity prediction.

\paragraph{Conditioning.}
The image and camera conditions enter the flow-matching network through
separate paths. A frozen DINOv2 ViT-L/14 \cite{oquab2023dinov2} encodes the reference image into
$18\times18$ 
patch features, which are combined with the corresponding
Pl\"ucker-ray embeddings to form $324$ conditioning tokens. These tokens
are prepended to $\mathbf Z_t$, and the complete sequence is processed
jointly by self-attention. 
Camera control is applied directly to $\mathbf Z_t$: the Pl\"ucker ray
map of each view is patch-embedded and average-pooled into a single
vector, which is added to all $16$ latent features of that view.

\paragraph{Training.}
Following the general strategy of noise-augmented
decoding~\cite{zheng2026rae}, we fine-tune the reconstruction model with
additive Gaussian perturbations applied to the latent scene features,
improving its robustness to latent perturbations. We then freeze the
adapted encoder and Gaussian decoder during flow-matching training. The
flow-matching network regresses the velocity using a mean-squared-error
objective~\cite{lipman2023flow}. Timesteps follow a logit-normal
distribution with a dimension-dependent shift, following
\cite{wei20263drae}. During training, the image condition is dropped with
probability $0.1$ to enable classifier-free guidance at sampling time.

We train the flow-matching network for $100{,}000$ steps using the full
reference image and then fine-tune it for a further $50{,}000$ steps
using random crops covering one quarter of the image area. Both training
periods use a per-GPU batch size of $32$ and a peak learning rate of
$1\times10^{-4}$ with $1{,}000$ warm-up steps. Each crop is resized to
the original conditioning resolution before tokenization, preserving
the $324$-token layout.

\paragraph{Sampling and Decoding.}
We sample with $50$ Euler steps using classifier-free guidance with a
scale of $1.5$. The generated latent scene features are denormalized and
combined with the camera references of the requested views to form the
compact scene representation $\mathbf S$, following
\cref{eq:scene_token_construction}. The frozen Gaussian decoder maps
$\mathbf S$ in a single forward pass to an explicit \gs{} scene, which
can subsequently be rendered from any viewpoint in real time.

\clearpage

\end{document}